\documentclass[letterpaper, 10 pt, conference]{ieeeconf}  

\IEEEoverridecommandlockouts                              
\usepackage{comment}
\usepackage{amsmath} 

\let\labelindent\relax
\usepackage{amsthm}
\usepackage{amssymb}  
\usepackage{xcolor}
\usepackage{cuted}
\usepackage{algorithm}
\usepackage{subcaption}
\usepackage{graphicx}
\usepackage{enumitem}
\usepackage{algpseudocode}
\usepackage{tikz}
\usetikzlibrary{decorations.pathreplacing,calc}
\newcommand{\tikzmark}[1]{\tikz[overlay,remember picture] \node (#1) {};}

\newcommand*{\AddNote}[4]{%
    \begin{tikzpicture}[overlay, remember picture]
        \draw [decoration={brace,amplitude=0.5em},decorate,ultra thick,black]
            ($(#3)!(#1.north)!($(#3)-(0,1)$)$) --  
            ($(#3)!(#2.south)!($(#3)-(0,1)$)$)
                node [align=center, text width=2.5cm, pos=0.5, anchor=west] {#4};
    \end{tikzpicture}
}%

\newtheoremstyle{mystyle}
  {}{}{}{}{\bfseries}{.}{.5em}{{\thmname{#1 }}{\thmnumber{#2}}{\thmnote{ (#3)}}}
\theoremstyle{mystyle}

\newtheorem{problem}{Problem}
\newtheorem{assumption}{Assumption}
\newtheorem{remark}{Remark}

\title{\LARGE \bf
Diagnosing and Augmenting Feature Representations \\ in Correctional Inverse Reinforcement Learning
}   
\author{In\^{e}s Louren\c{c}o, Andreea Bobu, Cristian R. Rojas, and Bo Wahlberg
\thanks{This work was supported by the Wallenberg AI, Autonomous Systems and Software Program (WASP), the Swedish Research Council Research Environment NewLEADS under contract 2016-06079, the Digital Futures project EXTREMUM, and Apple AI/ML Fellowship.}
\thanks{I. Lourenco, C. R. Rojas, and B. Wahlberg are with the Division of Decision and Control Systems, KTH Royal Institute of Technology, Stockholm, Sweden
        {\tt\small \{ineslo,crro,bo\}@kth.se}}%
\thanks{A. Bobu is with the Department of Electrical Engineering and Computer Sciences, University of California, Berkeley, Berkeley, CA, USA
        {\tt\small abobu@berkeley.edu}} %
}

\begin{document}

\maketitle
\thispagestyle{empty}
\pagestyle{empty}

\begin{abstract}

%


Robots have been increasingly better at doing tasks for humans by learning from their feedback, but still often suffer from model misalignment due to missing or incorrectly learned features.
When the features the robot needs to learn to perform its task are missing or do not generalize well to new settings, the robot will not be able to learn the task the human wants and, even worse, may learn a completely different and undesired behavior.
Prior work shows how the robot can detect when its representation is missing some feature and can, thus, ask the human to be taught about the new feature; however, these works do not differentiate between features that are completely missing and those that exist but do not generalize to new environments. In the latter case, the robot would detect misalignment and simply learn a new feature, leading to an arbitrarily growing feature representation that can, in turn, lead to spurious correlations and incorrect learning down the line.
In this work, we propose separating the two sources of misalignment: we propose a framework for determining whether a feature the robot needs is incorrectly learned and does not generalize to new environment setups \textit{vs.} is entirely missing from the robot's representation. Once we detect the source of error, we show how the human can initiate the realignment process for the model: if the feature is missing, we follow prior work for learning new features; however, if the feature exists but does not generalize, we use data augmentation to expand its training and, thus, complete the correction.
We demonstrate the proposed approach in experiments with a simulated 7DoF robot manipulator and physical human corrections.

\end{abstract}
\section{INTRODUCTION}

\begin{figure}[t]
 \centering
    \includegraphics[width=0.47\textwidth]{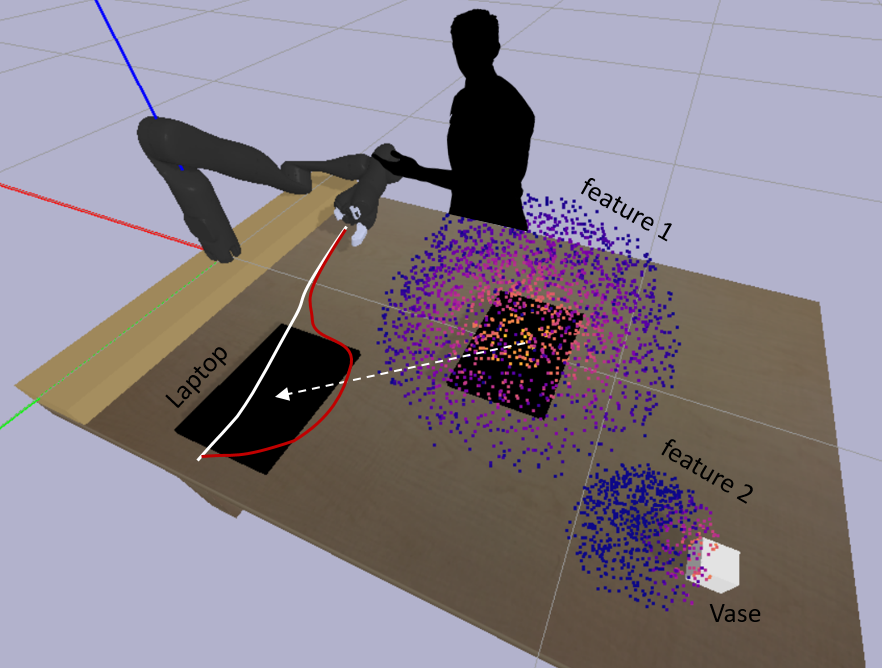}
    \caption{
    In a new environment, a laptop is encountered in a position the robot has never seen it. It is, therefore, not sure which (if any) of the features it learned during training are related to it. By analyzing the corrections the human applies to its trajectory in the vicinity of the laptop, the robot 
 \textit{i)} infers the relation between its learned features and the object (diagnosis), and \textit{ii)} adjusts the features to this new environment (augmentation).}
    \label{fig:general_view1}
\end{figure}

Communication is a key part of human life. When transmitting an idea to someone, we create \textit{representations}, or \textit{models}, of what we believe they are understanding or trying to make us understand. Communication flows naturally through this process until a behavior or response indicates that the model might be wrong. For instance, in Figure \ref{fig:general_view1} the human wants the robot to transport a cup of coffee while staying away from the laptop, but the robot has never seen the laptop in this position -- its model is different from the human's! At this point, an \textit{intervention} is typically conducted in the form of asking for an explanation or justification for the unexpected behavior. This intervention, despite not being constantly necessary, becomes crucial for the mutual understanding of the circumstances and the \textit{realignment} of the models. In our \textit{Human-Robot Interaction} (HRI) example, the human can explain to the robot how to correct misaligned features by performing an intervention in the form of a physical correction. 


The need for seamless and effortless interactions like this between humans and robots becomes fundamental as robots become increasingly more apt to perform tasks for and with humans. 
Misalignment happens when the robot's model of the human is incomplete, incorrect, or outdated, or when the human's model of the robot is similarly limited. It can lead to misunderstandings and ineffective interactions that are particularly undesirable when robots are expected to interact with humans in a natural and intuitive manner.
The interventions can take the form of, albeit not being limited to, demonstrations \cite{abbeel2004apprenticeship}, corrections \cite{bajcsy2017learning}, comparisons~\cite{christiano2017deep} and teleoperation \cite{javdani2015shared}.

Inverse reinforcement learning (IRL) is currently the most popular method for robots to estimate, from human behavior, the preferences and goals of those that surround them, typically in the form of a reward or cost function.
Traditional methods in IRL, however, assume that the human preferences are present in the robot's representation. If this assumption does not hold, the cost function the robot learns from the human's demonstrations and physical corrections is defective.

As in the previous example, in this work we consider that humans can communicate their preferences to robots by intervening on their trajectories by means of physical corrections, entering the field of physical Human-Robot Interaction (pHRI). To convey to the robot that it should transport the cup of coffee far away from the laptop, the human can push the robot's arm to the side to prevent passing above it. If the robot has a feature related to the distance to objects in its feature representation, it will understand the correction, update its model, and replan the trajectory to continue further away from the laptop. But if it does not -- or if the feature was not trained to generalize to this environment configuration -- it will assign a wrong feature to try to explain the correction, hindering the learning process.

We build upon previous work based on detecting misalignment by estimating the robot's confidence in human corrections, only updating its knowledge if it is confident that it understands the correction. One advantage of this approach is avoiding learning from noisy actions when the human is not optimal.
However, 
we argue that even when the human is optimal, the existence of misalignment associated with the robot's model can prevent the robot from understanding the correction. In that case, we propose that there is more information in the correction that the robot should take advantage of, in order to correct the misalignment. 

One well-known example of representation misalignment is the shift in features at inference vs. training time, due to the difficulty of models to generalize to different settings. This is commonly the case in IRL since interventions require human effort and therefore generalizing the model to all unexplored settings is challenging. Consider, for example, that there was a vase on top of the laptop during training. Now that the laptop is in a different position, should the learned features change with it, or are they instead related to the vase, which stayed in the same place? In other words, which objects did the features depend on? In this work, we propose a framework to identify incorrect features that do not generalize to new environments and completely missing features, where by estimating the relation between features and environment, the former can adapt to new settings. 

The main contributions of this work are:
\begin{itemize}
\item \textit{Misalignment diagnosis}  -- 
when a robot detects misalignment in its feature representation by not being able to understand human input, we design a framework that enables it to identify which features are misaligned, and disambiguate between incorrectly learned features that do not generalize to new environment setups and entirely missing features from the robot's model.
\item \textit{Misalignment correction} -- we propose a framework that enables robots to realign the misaligned features by using data augmentation to generalize them to the new task, and the missing features by asking the human for new data about the feature.
\item We evaluate the two methods on a 7DoF simulated robotic arm that aims to align its preferences to the human's from the physical torques applied during its trajectory.

\end{itemize}

The remainder of this paper is organized as follows: Section \ref{ssec:lit_review} gives an overview of related literature in the field, Section \ref{sec:problem_form} formulates an IRL framework to detect confidence in the human input, and Section \ref{sec:generalizing_feats} presents and discusses our proposed methods to diagnose and correct misaligned features. Finally, Section \ref{sec:experiments} evaluates how the proposed framework works in a 7-DoF robotic manipulator, and Section \ref{sec:conclusions} concludes with a discussion of some of the advantages and open problems of our framework, as well as suggestions for future research directions.

\section{RELATED WORK}
\label{ssec:lit_review}

\textbf{Inverse reinforcement learning}
is a popular framework for learning cost functions from human demonstrations \cite{ng2000algorithms, ziebart2008maximum}, which considers the human as a utility-driven agent that chooses actions with respect to an internal cost function. As an approach to Inverse Optimal Control that does not require a model of the environment, it is particularly useful when the cost functions are difficult or impractical to manually design.

\textbf{Learning from corrections} is another way of learning from human input that can be a good complement and advantageous in many situations where real-time and task-specific learning is needed \cite{lourencco2021cooperative, lourencco2022teacher}. Methods to incorporate corrections in real-time to align robot and human preferences have been shown to improve performance and adaptability for HRI \cite{bajcsy2017learning, bajcsy2018learning, jain2015learning, gutierrez2018incremental}.

\textbf{Uncertainty in robot learning} can be incorporated into the representations learned by maintaining a probability distribution over what the cost functions might be \cite{hadfield2017inverse,ramachandran2007bayesian}. In
\cite{losey2018including} the authors proposed using a Kalman filter to reason over the uncertainty of the estimated human preferences from physical corrections.
However, even by keeping track of uncertainty, these works still assume that the human preferences lie in the robot’s representation.
In \cite{bobu2018learning} and \cite{bobu2020quantifying}, a formulation is proposed where the robot's representation does not necessarily have to fully capture the human's underlying preferences. The authors propose learning the features proportionally to the robot's confidence in the human input, assuming low confidence to be a result of noisy or suboptimal human actions and leaving for future work expanding the robot's feature representation. In \cite{bobu2021feature}, the authors assume misalignment to be the result of missing features in the robot representation and solve it by querying the human for new input. 

\textbf{Generalization to new environments} is a challenge in IRL which typically requires training from demonstrations rich enough to incorporate a wide variety of states of the environment. Methods like transfer learning \cite{weiss2016survey} exist to adapt the learning in real-time or when the cost of retraining the model is prohibitively high.
Our proposed framework fine-tunes the cost function to unexplored environments, by adjusting its features according to the changes in the environment.

\section{PROBLEM FORMULATION}
\label{sec:problem_form}

In this section we formalize the IRL framework used to teach a robot a cost function from offline demonstrations, how this cost function can be continuously improved from online human corrections, and how misalignment in the cost functions of the human and of the robot can be detected online by computing the robot's confidence in the human input.

\subsection{Offline cost learning from demonstrations}


IRL is a technique used to infer a cost function $C$ from a given set of demonstrations $\mathcal{D} =\{\xi_1, \ldots, \xi_D\}$. To make the problem tractable, $C$ is typically parametrized by a vector $\theta \in \Theta$, denoting the preferences of the human for performing the task, and the aim of the robot is to estimate these parameters. If $\theta$ was known by the robot, the problem could be treated as a Markov Decision Process (MDP). However, uncertainty over $\theta$ turns it into a Partially-Observable MDP (POMDP) formulation, with $\theta$ as a hidden part of the state. In this setting, the human actions are thus observations about $\theta$ under some observation model $P(\xi|\theta)$.

A popular decision-making model for human behavior is the Boltzmann model, which considers humans as noisily-optimal agents that typically choose control inputs that approximately minimize their cost \cite{von2007theory, baker2007goal}. According to this model, the probability of seeing a demonstration depends exponentially on its cost $C_\theta(\xi)$: 
\begin{equation}
    P(\xi|\theta)=\frac{e^{-C_\theta(\xi)}}{\int e^{-C_\theta(\bar{\xi})} d \bar{\xi}}.
    \label{eq:obs_model}
\end{equation}
By assuming $\{\xi_1, \ldots, \xi_M\}$ \textit{i.i.d.} and using maximum likehood with the Monte Carlo method, the estimated parameters $\hat{\theta}$ are then the ones that maximize the probability of seeing the demonstrations:
\begin{equation}
    \hat{\theta} = \arg \max_\theta \mathcal{L}(\theta) \approx D \log \sum_{i=1}^D e^{-C_\theta(\xi_i)} - \sum_{i=1}^D C_\theta(\xi_i).
    \label{eq:IRL_theta}
\end{equation}

Once an estimate of the cost function has been computed from training demonstrations, the robot then uses it to perform its task accordingly. 

\subsection{Online cost update from corrections}

In order to enable the learning process of the robot to continuously adapt and adjust in real-time to different settings, work has been developed to apt it to learn from other sources of human input, such as physical corrections~\cite{bajcsy2017learning}. 

Formally, the problem can be formulated as a dynamical system $\dot{x} = f(x,u_R+u_H)$, where $x$ is the state of the robot including its position and velocity, $u_R$ is its action (e.g. the torque applied at the joints) and $u_H$ the external torque applied by the human. 
There is a true objective function $C_\theta$ known by the human but not by the robot. 
One of the conventional models to approximate the infinite-dimensional space of possible cost functions is using basis functions \cite{abbeel2004apprenticeship, ziebart2008maximum}, based on which the cost can be written as a linear combination of features of the state, $C_\theta(x)=\theta^T\phi(x)$, which can be arbitrary mappings $\phi \colon \mathbb{R}^d \rightarrow [0,1]$.
While the relevant set of features $\phi$ is known by the robot from previous training, the weights, $\theta$, denoting the preferences of the human for performing the task, must be adjusted to the task at hand. We use this model henceforth throughout the paper. 
The correctional HRI framework is described next.\\

\paragraph{The robot acts} The robot uses its estimate of $\hat{\theta}$ obtained from \eqref{eq:IRL_theta} to compute a trajectory $\xi_R=\{x_t\}_{t=0}^T$ that minimizes its current cost function $C_\theta$, and finds the control inputs $\{u_{R,t}\}_{t=0}^{T}$ to follow it. The cost of a trajectory is written as $C_\theta(\xi)=\theta^T\Phi(\xi)$, where $\Phi(\xi) = \sum_{x\in\xi} \phi(x)$ is the sum of the features $\phi$ along the trajectory $\xi$. 
The policy optimization scheme is given by
\begin{equation}
    \xi_R = \arg \min_\xi \hat{\theta}^{T} \Phi(\xi).
    \label{eq:opt_trajectory}
\end{equation}


\paragraph{The human corrects} If the robot follows a trajectory that does not seem correct to a noisily-rational human, they can, at a given time, choose to induce a joint torque $u_H$ to deform the original trajectory to $\xi_H$. The deformed trajectory is given by $\xi_H = \xi_R + \mu A^{-1}U_H$ \cite{bajcsy2017learning}, where $\mu$ determines the magnitude of the deformation, the matrix $A \in \mathbb{R}^{T\times T}$ determines its shape, and $U_H=u_H$ at the moment of the correction and is 0 otherwise. The correction is done with the goal of minimizing the cost, while also minimizing human effort $u_H$. Hence, the observation model from \eqref{eq:obs_model} can be rewritten as 
\begin{equation}
    P(u_H|\xi_R;\theta) = \frac{e^{-(\theta^T\Phi(\xi_H)+\lambda||u_H||^2)}}{\int e^{-(\theta^T\Phi(\bar{\xi}_H)+\lambda||\bar{u}_H||^2)} d\bar{u}_H },
\label{eq:human_action_cost}
\end{equation}
where $\lambda$ symbolizes the trade-off between cost and human effort.

\paragraph{The robot updates its knowledge} 
By computing the difference between the sum of the features of the original and deformed trajectories, the robot updates $\hat{\theta}$ as 
\begin{equation}
    \hat{\theta} \leftarrow \hat{\theta} - \alpha(\Phi(\xi_H)-\Phi(\xi_R)),
    \label{eq:theta_update_IRL}
\end{equation}
where $\alpha\geq 0$. More details on the derivations can be found in \cite{bajcsy2017learning}, but the intuitive interpretation is that the feature weights are updated based on the direction of the change of the feature values between the original and deformed trajectories. If the deformed trajectory passes further away from an object, the weights of the corresponding distance-to-object feature will increase.

\subsection{Online misalignment detection} 
\label{ssec:mis_detect}

When a robot acts as described above, it takes the risk of naively trusting every human action, thus possibly learning from incorrect information. As mentioned in Section \ref{ssec:lit_review}, methods exist in the literature for keeping track of uncertainty, but a solution to the problem of whether the desired features exist in the robot's feature representation was, until recently, missing. 

In \cite{bobu2018learning}, a method was presented to study when human corrections cannot be explained by the robot's model. The misalignment detection problem is tackled by adding a rationality coefficient $\beta\in[0,\infty)$ to \eqref{eq:human_action_cost}, to account for uncertainty in how the human picks $u_H$. The computation of the two inference parameters, $\beta$ and $\theta$, can be separated into two parts and consists of analyzing how efficiently the feature values $\Phi(\xi_H)$ of the deformed trajectory can be explained by each of the robot's features.
In a nutshell, the robot computes $\Phi(\xi_H)$ and searches for corrections, $u_H^*$, from the original trajectory, that could have achieved the same feature value change. The smallest correction would be the one that the human would have performed if it was performing the correction because of dissatisfaction with that robot feature. This is formulated as the following constrained optimization problem:
\begin{equation}
\begin{aligned}
    \min_{\bar{u}_H} & \quad ||\bar{u}_H||^2 \\
    s.t. & \quad \Phi(\xi_R + \mu A^{-1}\bar{u}_H) - \Phi(\xi_H) = 0.
\end{aligned}
\label{eq:optimal_u_star}
\end{equation}
Then, by computing how far the optimal correction $u_H^*$ is from the actual correction $u_H$ received, the robot estimates its confidence $\hat{\beta}$ in the human input $u_H$: 

 \begin{equation}
    \hat{\beta} = \frac{k}{2\lambda(||u_H||^2-||u_H^*||^2)},
    \label{eq:beta}
\end{equation}
where $k$ is the dimension of the action space.

Once we have an estimate of the confidence $\hat{\beta}$, we can compute the new posterior estimate of the human's cost function $p(\theta|\Phi(\xi_H),\hat{\beta})$, and from it obtain the update rule
\begin{equation}
    \hat{\theta} \leftarrow \hat{\theta}-\alpha \frac{\Gamma\left(\Phi(\xi_H), 1\right)}{\Gamma\left(\Phi(\xi_H), 1\right)+\Gamma\left(\Phi(\xi_H), 0\right)}\left(\Phi(\xi_H)-\Phi\left(\xi_R\right)\right);
    \label{eq:new_theta_update}
\end{equation}
where
\begin{equation}
\Gamma\left(\Phi(\xi_H),i\right)=P(E=i \mid \hat{\beta}) P\left(\Phi(\xi_H) \mid \theta, E=i\right),
\label{eq:E_beta}
\end{equation}
and $E$ is a proxy variable for $\hat{\beta}$. The interested reader is referred to \cite{bobu2018learning} for more details. It is, however, relevant to point out that if the possibility of the representations being misaligned is not taken into consideration, \eqref{eq:new_theta_update} simplifies to~\eqref{eq:theta_update_IRL}.

Once $\hat{\theta}$ is updated, the robot goes back to the task at hand.\\

In summary, the robot updates the parameters $\hat{\theta}$ of the cost function proportionally to its confidence $\hat{\beta}$ on the human correction. If some of its features can explain the correction, such as the human pushing the robot arm in the opposite direction of the laptop, $\hat{\beta}$ is large and the weight $\hat{\theta}$ of the distance-to-the-laptop feature is increased to represent that the human cares more about the distance to the laptop than previously estimated. If, on the other hand, $\hat{\beta}$ is small, such as in the case that the robot does not know about the distance-to-laptop feature, no values of $\hat{\theta}$ can make the cost function explain the correction; so $\hat{\theta}$ is not updated much. This confidence-based model update method is presented in Algorithm~\ref{alg:mis_detection}.


\begin{algorithm}[t]
\caption{Misalignment detection}
\begin{algorithmic} 
\Require $\phi_i$ for $i=1,\ldots,M$
\Require $\mathbf{\xi}_R = \arg \min_\xi \hat{\theta}^{T} \Phi(\xi)$, for initial $ \hat{\theta}$.
\While{goal not reached}
\If{$u_H \neq \mathbf{0}$}
\State $ \mathbf{\xi}_H=\mathbf{\xi}_R+\mu A^{-1} \tilde{{u}}_H$ \tikzmark{top3} 
\State $ u_H^*$ computed from \eqref{eq:optimal_u_star} \qquad \tikzmark{right3} 
\State $ \hat{\beta}=\frac{k}{2 \lambda\left(\left\|u_H\right\|^2-\left\|u_H^*\right\|^2\right)}$ \tikzmark{bottom3}
\If{$\hat{\beta}$ small} 
\State \textit{diagnose\_and\_correct\_misalignment()}, in Alg. \ref{alg:mis_id_correction}
\State Recompute $u_H^*$ and $\hat{\beta}$
\EndIf
\State $\hat{\theta} \leftarrow \hat{\theta}-\alpha \frac{\Gamma\left(\Phi(\xi_H),1\right)}{\Gamma\left(\Phi(\xi_H), 1\right)+\Gamma\left(\Phi(\xi_H), 0\right)}\left(\Phi(\xi_H)-\Phi\left(\mathbf{\xi}_R\right)\right) .$
\State $\mathbf{\xi}_R = \arg \min_\xi \hat{\theta}^{T} \Phi(\xi)$
\EndIf
\EndWhile
\AddNote{top3}{bottom3}{right3}{\textit{Detect misalignment}}
\end{algorithmic}
\label{alg:mis_detection}
\end{algorithm}

\section{MODEL ALIGNMENT \\THROUGH FEATURE GENERALIZATION}
\label{sec:generalizing_feats}

In this section, we present how the cost function of the robot can be improved based on the human corrections, even when $\hat{\beta}$ is low. Recall that a low value of $\hat{\beta}$ means that there is no $\theta$ that can make the cost function $C_\theta (x)=\sum_{i=1}^{M} \theta_i \sum_{t=0}^T \phi_i\left(x^t\right)$ optimal in the light of the human corrections. Hence, the misalignment of the cost function has to be a result of a misaligned feature representation, $\phi_R= \{\phi_1,\ldots,\phi_{M}\}$; that is, $\phi_R \neq \tilde{\phi}_H$. It must be either the case that a feature is missing, $\exists \; \tilde{\phi}_i \in \tilde{\phi}_H \text{ s.t. } \phi_i \notin \phi_R$, or that at least one feature is incorrect, $\exists \; \tilde{\phi}_i \in \tilde{\phi}_H \text{ s.t. } \phi_i \in \phi_R \text{, but } \tilde{\phi}_i \neq \phi_i$. 

The only existing relevant approach in the current literature to overcome representation misalignment is to query the human for more data, thus assuming misalignment stem from the first case, which is an incomplete feature representation \cite{bobu2021feature} ($\tilde{\phi}_i \notin \phi_R$). However, if the misaligned features actually exist in the robot's representation but were incorrectly learned and do not generalize to new environments ($\tilde{\phi}_i \neq \phi_i$), this assumption can lead to spurious correlations and incorrect learning down the line.
To overcome this problem, in this section we propose a framework to diagnose model misalignment by differentiating between the two sources of misalignment. 

First, one common reason for $\tilde{\phi}_H \neq \phi_R$ is that $\phi_R$ consists of features learned for a specific training environment, which are not generalizable for new environments that the robot acts in, and, thus, become misaligned with the human's features $\tilde{\phi}_H$ in the new environment. The challenge of generalizing features to new settings stems from the fact that the features robots learn are a function of all the objects, and their respective positions, present in the environment at the moment of training, which makes the generalization problem highly dependent on large training datasets with different object positions.


Inspired by recent works on learning in real-time, in this section we propose an approach to instead address the generalization problem online as the robot performs the task, simply from human corrections. 
We tackle questions like: if the robot learns from demonstrations given when a vase is on top of a laptop, when one of the objects moves, should the feature learned from the demonstrations move as well?
When acting in different settings where the objects the robot has to interact with are in new positions, the problem becomes understanding i) how the learned features are related to the various objects, and ii) how they should be aligned when the objects move. We divide this two-fold problem into two main goals, which are each tackled in each of the next subsections, and conclude the section with some final remarks.

\begin{problem} [Misalignment diagnosis] Once misalignment has been detected from \eqref{eq:beta}, how can we identify if it stems from existing features that have not generalized to the new environment or completely missing features?
\end{problem}

\begin{problem}[Misalignment correction] After diagnosing the misaligned features, how can we align the representations by translating them to the new environment? 
\end{problem}

\subsection{Diagnosing misaligned features}
\label{ssec:id_framework}




To make explicit the dependency of the state $x$ on the surrounding environment, we define it as in \cite{bobu2018learning} according to the position of the robot joints, and of the objects in the environment. In this way, $x = \{R, o^1, \ldots, o^N\}$.
Since the robot only has access to features $ \phi = \{\phi_1,\ldots,\phi_M\}$ trained in these states, we write them as  $\phi(R,o^1,\ldots,o^N)$. Due to the inexistence of a map between features and objects during training, the question then becomes how the robot can estimate a changed feature representation $\tilde{\phi}(R, \tilde{o}^1, \ldots,\tilde{o}^N)$, where the features are now adapted to the testing setting. Here, $\tilde{o}^i$ represents the new position of object $i$, and $\tilde{R}$ the new position of the robot. 

When tasked with computing an optimal  trajectory in the new environment, the robot starts by computing a value $\Delta$ for how much each object moved, $\Delta_i = o^{i}-\tilde{o}^i$, for $i=1,\ldots,N$.
Initially, the robot does not know how the features should be translated so it plans the trajectory according to \eqref{eq:opt_trajectory}, where $\Phi$ is the sum of the values of the trained features $\phi(R, o^1,\ldots,o^N)$. 
%
If no corrections are received while it follows the trajectories throughout the environment, the shifted objects did not actually have an impact on the features so the robot can correctly perform the tasks.
If corrections are received but $\hat{\beta}$, computed according to \eqref{eq:beta}, is large, then the features are still correctly represented and the robot just has to update their importance $\hat{theta}$ for that task as in \eqref{eq:new_theta_update}.
If, on the other hand, $\hat{\beta}$ is small, no $\hat{\theta}$ can explain the corrections and therefore the feature representation $\phi(R, o^1,\ldots,o^N)$ has to be corrected to $\tilde{\phi}(\tilde{R}, \tilde{o}^1,\ldots,\tilde{o}^N)$.

\begin{figure}[t]
         \centering
         \begin{subfigure}[t]{0.225\textwidth}
            \centering
            \includegraphics[width=\textwidth]{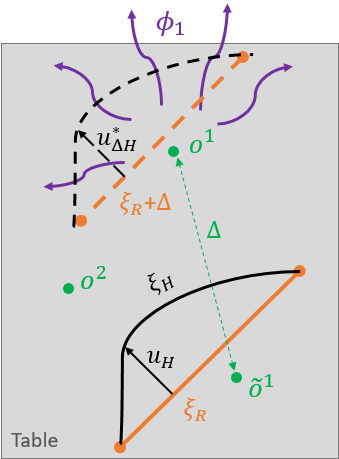}
            \caption{Large $\hat{\beta}$. The shifted trajectory $\xi_R+\Delta$ passes over a significant feature $\phi_1$, so the optimal correction $u_H^*$ would be large and thus similar to $u_H$. The feature $\phi_1$ is related to the moved object $o^1$.}
            \label{fig:shift1}
         \end{subfigure}
         \hfill
         \begin{subfigure}[t]{0.23\textwidth}
             \centering
             \includegraphics[width=\textwidth]{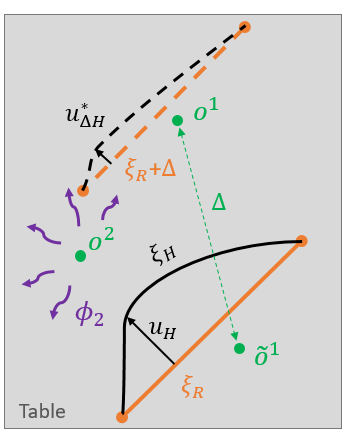}
             \caption{Small $\hat{\beta}$. In this case, the feature $\phi_2$ does not have a significant influence on the shifted trajectory $\xi_R+\Delta$, so $u_H^*$ is quite small, and therefore different from $u_H$. The feature does not depend on $o^1$.}
             \label{fig:shift2}
         \end{subfigure}
            \caption{Schematic illustrations of how the misaligned features can be diagnosed when the object $o^1$ is moved by $\Delta$.}
            \label{fig:scheme_featureId}
    \end{figure}

To compute, the misaligned features from the correction, we use a similar idea to that in Section \ref{ssec:mis_detect}: we compute a confidence in how well each feature can explain the correction, but now taking into consideration that some features might have to be changed according to the moved objects. We denote this new confidence $\beta_\Delta$, and compute it according to
\begin{equation}
    \hat{\beta}_\Delta = \frac{k}{2\lambda(||u_H||^2-||u_{H\Delta}^*||^2)}.
    \label{eq:beta_delta}
\end{equation}
The parameters are as in Section \ref{ssec:mis_detect}, but the comparison is now done between the human correction gotten, and the one that would have been optimal for that feature if it depended on the shifted object. 

The shifted optimal correction, $u_{H\Delta}^*$, is now computed by shifting the original and deformed trajectories by $\Delta$ as if the object was still in the same position as in training, and in that state computing the corresponding values for the trained features. In order words, we need to compute the feature values in a new state that was unexplored during training, but whose values might be the same, if the feature belongs to the moved object, as those in a known training state. This can be formulated as the constrained optimization problem
\begin{equation}
\begin{aligned}
    \min_{\bar{u}_H} & \quad ||\bar{u}_H||^2 \\
    s.t. & \quad \Phi(\xi_R + \Delta + \mu A^{-1}\bar{u}_H ) - \Phi(\xi_H + \Delta) = 0.
\end{aligned}
\label{eq:h_h_star_delta}
\end{equation}
Figure \ref{fig:scheme_featureId} schematically illustrates how \eqref{eq:h_h_star_delta} computes what the optimal correction for the different features would have been, if the object had not changed position. 
Hence, $\hat{\beta}_\Delta$ represents the confidence of the robot on the ability for the features to explain the correction if they were shifted with the changed object. 

An important condition must, however, be fulfilled for this to be true.

\begin{assumption}
The corrections are performed by taking into consideration one feature at a time.
    \label{ass:one_corr}
\end{assumption}
    This assumption determines our ability to consider that the contribution of each feature can be individually compared with the correction. This is, however, a common assumption in the literature, including the methods in Section \ref{sec:problem_form}, and has been shown to be reasonable. For example, \cite{bajcsy2018learning} studied the advantages of correcting a IRL robot considering one feature at a time.

\algnewcommand{\algorithmicgoto}{\textbf{go to}}%
\algnewcommand{\Goto}[1]{\algorithmicgoto~\ref{#1}}%
\begin{algorithm}[t]
\caption{Misalignment diagnosis and correction}
\begin{algorithmic}[1]
\Require $\xi_R, u_H, \xi_H, \phi(R,o^1,\ldots,o^N), \tilde{R}, \tilde{o}^{1},\ldots,\tilde{o}^N$ 
\State $\Delta_i=o^i-\tilde{o}^{i}$ 
\State $missing\_feature = 1$
\For{$\phi_1,\ldots,\phi_M$} 
\State Compute $\Phi(\xi_R+\Delta)$, $\Phi(\xi_H+\Delta)$ \tikzmark{top1} \tikzmark{right1} \; \tikzmark{right2}
\State Solve \eqref{eq:h_h_star_delta} to get $u_{H\Delta}^*$
\State Solve \eqref{eq:beta_delta} to get $\hat{\beta}_\Delta$ \tikzmark{bottom1}  
\If{$\hat{\beta}_\Delta$ large} \Comment{$\phi_k$ is a function of $o^{i}$}
\State $inverse\_kinematics(j_0,j_{J}+\Delta)$ \tikzmark{top2} 
\State Correct $\phi$ to $\tilde{\phi}$ according to \eqref{eq:correct_feats} 
\State $missing\_feature = 0$ \tikzmark{bottom2}
\EndIf
\EndFor
\If{$missing\_feature$}
\State{Query human for demonstrations for $\tilde{\phi}_{M+1}$ }
\EndIf
\AddNote{top1}{bottom1}{right1}{\textit{Diagnose misalignment}}
\AddNote{top2}{bottom2}{right2}{\textit{Correct misalignment}}
\end{algorithmic}
\label{alg:mis_id_correction}
\end{algorithm}

\subsection{Feature alignment}
\label{ssec:corr_framework} 

In the previous section we computed the confidence $\hat{\beta}_\Delta$ of the robot on each feature $\phi$. If this confidence is small for all the features, none of them can explain the correction received in the new environment. In this case, the misalignment has to derive from a missing feature, for which the robot was not trained. To correct this misalignment the robot proceeds by asking the human for data to learn the new feature. Details on how the cost function can be augmented by learning new features from neural networks can be found in \cite{bobu2021feature}.

If, on the other hand, the confidence $\hat{\beta}_\Delta$ is large for some features, to correct the misalignment the robot must align them to the new setting. 
This can be done according to
    \begin{equation}
    \begin{aligned}
    \tilde{\phi}_i(\tilde{R}, \tilde{o}^1,\ldots,\tilde{o}^N) \leftarrow     \phi_i(R, o^1-\Delta_1,\ldots,o^N-\Delta_N),
    \label{eq:correct_feats}
    \end{aligned}
    \end{equation}
where $\tilde{\phi}_i$ is the updated feature $\phi_i$, $\tilde{o}^i$ is the new position of the shifted object, and $\tilde{R}$ is the new position of the robot joints. 
Since the features depend not only on the object locations but also on the joint positions of the robot, to align the features we also need to compute the shift in the robot joint positions.
Let us consider a robotic arm with $J$ joints $j_1,\ldots,j_J$. The base joint $j_0$ will remain the same if the robot does not move, but the $xyz$ position of the end-effector joint, $j_J$, has to be moved by $\Delta$, and the  be the same.
Then, for joints $j_1,\ldots,j_{J-1}$, we can use standard inverse kinematics methods to compute the joint values that lead to the required end-effector poses. \\

The complete diagnosis and alignment framework is summarized in Algorithm \ref{alg:mis_id_correction}, for the misaligned object $o^i$. This procedure can be repeated for others. Once the features are aligned, the robot completes Algorithm \ref{alg:mis_id_correction} and returns to Algorithm \ref{alg:mis_detection}.
The weights of the newly aligned features still need, however, to be adjusted to the new environment. So from the same correction $u_H$, as before, the robot estimates the new $u_H^*$ and $\hat{\beta}$.
Based on these it then updates the weights $\hat{\theta}$ according to \eqref{eq:new_theta_update}, and can resume its trajectory according to \eqref{eq:opt_trajectory}, now for $\Phi$ being the sum of the values of the aligned features $\tilde{\phi}(\tilde{R}, \tilde{o}^1,\ldots,\tilde{o}^N)$ from \eqref{eq:correct_feats}.






\subsection{Remarks} 
\label{ssec:framework_remarks}
Let us now discuss some details about the framework.

\textbf{Multi-object dependent features:} The proposed alignment framework presented in the previous subsections can be directly applied to features that depend on multiple objects, including concepts such as \textit{distance between objects}, \textit{above}, \textit{near}, \textit{aligned}, etc. (more examples can be found in \cite{bobu2022learning}). If one of the objects is moved in the inference setting, like in the case of the vase that was \textit{above} the laptop during training but moved, the \textit{above} feature will not be relevant in the new setting so the human will not correct the robot trajectory near the moved vase object. The representation is therefore not altered. Further, in the position where the objects were during training, the robot will now be corrected to assign a small weight $\theta$ to the feature.

\textbf{New inference objects:}
Despite having been presented for the case where objects are changed between training and testing settings, the framework can also be directly applied to settings where new objects are present at testing time -- as long as they are the same, or behavior-invariant versions (that is, the distinguishing characteristics, such as color, do not change how the object should be interacted with) -- of the training objects. Both $\Delta$ and $\beta_\Delta$ are computed as before, the only difference is that instead of altering feature $\phi_i$ as in \eqref{eq:correct_feats}, a new one (or more), $\phi_{M+1}$, is added: 
        \begin{equation}
    \begin{aligned}
    \tilde{\phi}_{M+1}(\tilde{R}, \tilde{o}^1,\ldots, \tilde{o}^N)
    = \phi_i(R, o^1-\Delta_1,\ldots,o^N-\Delta_N).
    \end{aligned}
    \end{equation}

\textbf{Object uncertainty:}
    This framework assumed that the robot knows which objects have changed position. If this is not the case $\Delta$ is not uniquely known but the procedure can still be repeated for multiple $\Delta$s corresponding to the distance to each of the training objects until the correct one is found. From $\hat{\beta}_\Delta$ the robot evaluates if any of the features for the training objects apply to the new object. For example, if a mobile phone is seen at testing but not training, the robot could associate it with technology and therefore act as it would with the computer. This assumption could then be confirmed or denied by the human corrections received. This can be extended by computing a probability $P(\tilde{o}^{N+1}=o^i)$ of the new observations being behavior-invariant versions of the known ones.

\textbf{Small and large $\hat{\beta}$:}
Throughout the paper we have used the concepts of small and large to define the confidences $\hat{\beta}$ and $\hat{\beta}_\Delta$, and therefore as a basis to define misalignment. Despite intuitively sound, this ends up being a threshold parameter that has to be tuned offline; and which can be represented by the binary variable $E=\{0,1\}$ defined in \eqref{eq:E_beta} as being a proxy to $\hat{\beta}$. In \cite{bobu2020quantifying}, 
this threshold is computed from user experiments using the Bayesian framework $P(E \mid \hat{\beta}) \propto P(\hat{\beta} \mid E) P(E)$, by fitting the distributions from controlled user interaction samples.

\textbf{Alignment policy:}
A related question to the tuning of the confidence threshold is what policy the robot should follow when it comes to deciding how many features it should assign to the moved objects. We did not go into details about this but argue that the policy can be flexible and adjustable to the task at hand. The robot might not mind the risk of making wrong assumptions about the features and being corrected, or might consider safety as a critical aspect of the task and thus use a higher confidence threshold.

\section{EXPERIMENTS}
\label{sec:experiments}

We evaluate our framework in a 7-DoF simulated JACO robotic manipulator implemented in Pybullet in a 1.80 GHz CPU. We added the ability for humans to correct the robotic arm trajectory by applying a torque to the robot using the cursor. Corrections to the real robot have been applied in a similar way \cite{bobu2020quantifying}.
We will study the behavior of the robot in an environment with two objects:  $o^1$, a black rectangle representing a \textit{laptop}, and $o^2$, a white cube representing a \textit{vase}; and two features: $\phi_1:$\textit{ distance-to-laptop}, a large point cloud in the center; and $\phi_2 :$\textit{ distance-to-vase}, a smaller point cloud on the bottom left corner. The feature distances are computed from the position of the end-effector of the robot to the center of the object. Note, however, that each feature $\phi_i(R, o^1,o^2), i=1,2$ is learned as a function of the robot position and both objects in the environment, meaning that the robot does not know how each feature is related to each of the objects individually.

At inference time, the robot needs to transport a cup of coffee across the table, in a new environment where, for example, the laptop is in a different position $\tilde{o}^1$. 
Using TrajOpt \cite{schulman2013finding}, the optimal trajectory $\xi_R$ is computed and followed according to \eqref{eq:opt_trajectory} based on the previously trained features, since the robot does not know which, (if any) features correspond to the moved object. 
Let us consider the case where the cup should be transported in the vicinity of the position of the vase. In this new environment, the human prefers the robot to keep a bigger distance from it than during training, so it slightly pushes the arm away from the vase to change the trajectory of the robot.
The robot follows Algorithm \ref{alg:mis_detection} and, through $\hat{\beta}$ from \eqref{eq:beta}, computes that its current features (in particular, the distance-to-vase) can explain the correction, so increases the weight $\theta$ of this feature translating that it learned to stay further away from the vase. In the rest of the section, we analyze how our proposed framework performs when, on the other hand, $\hat{\beta}$ is small so the correction cannot be explained by any of the existing features.

\begin{figure}[t]
 \centering
    \includegraphics[width=0.9\linewidth]{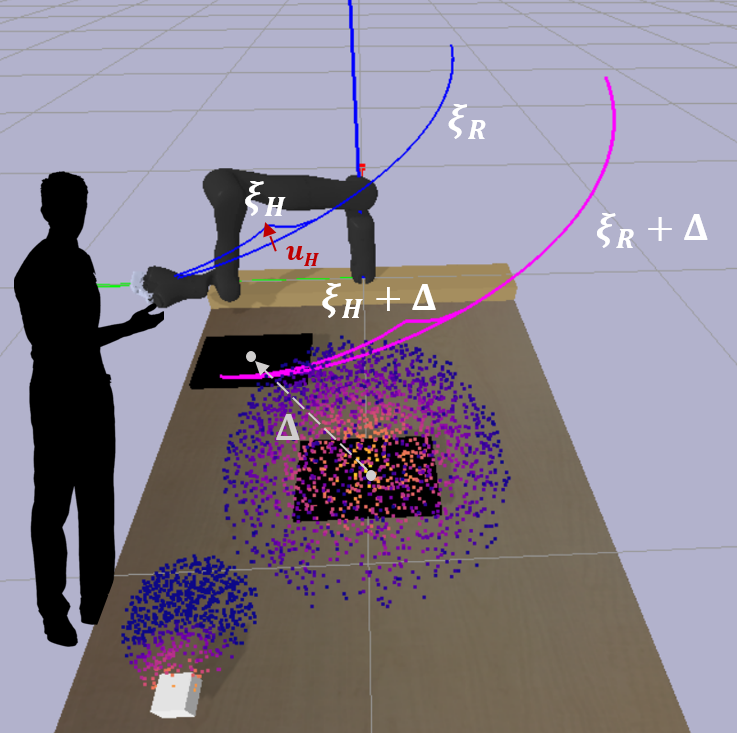}
    \caption{If the human correction $u_H$ cannot be explained by the current features $\phi$ of the robot, it translates the original and deformed trajectories $\xi_R$ and $\xi_H$ to where the object moved from and computes how well each of the features in that state could explain the received correction.}
    \label{fig:delta_traj}
\end{figure}

\begin{figure}[t]
 \centering
    \includegraphics[width=0.9\linewidth]{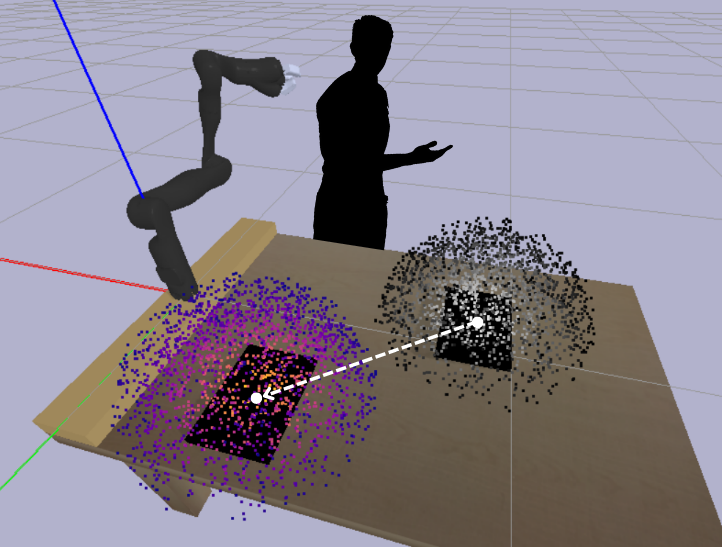}
    \caption{Once a misaligned feature is identified, the robot can successfully correct it to a new state according to the new object position.} 
    \label{fig:correct_misalignment}
\end{figure}

Let us now analyze the case where the trajectory passes in the vicinity of the new laptop position, illustrated in Figure \ref{fig:delta_traj}.
Since, unlike the robot, the human knows that the distance-to-laptop feature should be taken into account when performing a trajectory near the laptop, independently of its position, it applies a torque $u_H$ to correct the trajectory to $\xi_H$ to pass further away from it. 
Since none of its current features were trained for those states, they cannot explain the correction so the robot follows Algorithm \ref{alg:mis_id_correction} to estimate $\hat{\beta}_\Delta$ as in \eqref{eq:beta_delta}. It computes what the values $\Phi$ for the two trained features would be for the training position of the laptop, $\xi_R+\Delta$. In this case, $\Phi(\xi_R)=\Phi(\xi_H)=[0,0]^T$, $\Phi(\xi_R+\Delta)=[20.13,0]^T$ and $\Phi(\xi_H+\Delta)=[18.6,0]^T$. From \eqref{eq:h_h_star_delta}, it then computes the optimal correction $u_H^*$ that would have been applied to that shifted trajectory according to each of the features. 
The second elements are always zero because the distance-to-vase feature is far away from all four trajectories. Therefore, $u_H^*=0$ since the cost of that feature is already minimized, and the human has a preference for smaller corrections that minimize their effort \eqref{eq:human_action_cost}.
For the distance-to-laptop feature, on the other hand, the cost of the shifted deformed trajectory is decreased, resulting in a $u_H^*>0$, and equal to $u_H$.
Therefore, from \eqref{eq:beta_delta} we get a small $\hat{\beta}_\Delta$ for the former feature, but a large one for the latter, showing that the distance-to-laptop feature explains the correction performed near the new laptop position. In other words, that this feature is related to the object that moved.

Once the robot learns that the distance-to-laptop feature corresponds to the object moved, we apply the framework from Section \ref{ssec:corr_framework} to correct the feature values to the new setting, $\tilde{\phi}(\tilde{R},\tilde{o}^1,o^2)$. Figure \ref{fig:correct_misalignment} shows how the feature values change in the vicinity of the new (in purple) and previous (in grey) positions of the object.

Finally, the next time that the robot has to perform a task near the new or old positions of the object, it will already know how to so successfully. 

\section{CONCLUSIONS}
\label{sec:conclusions}
In this work we argued that robots' adaptation capabilities should go beyond detecting when their representations are misaligned with those of humans. We proposed a framework that enabled them to diagnose the causes of the misalignment -- by distinguishing between misalignment caused by incorrect features that do not generalize to new environments, and completely missing ones -- and a method to correct the misalignment in each of these scenarios.
While the latter cause of misalignment can be addressed by querying the human for more feature data, solving the former required estimating which objects the different learned features are related to. 
Our framework has the advantage of being applicable while the robot perfoms a task in real-time, leveraging information from physical human corrections.
In simulations we showed that a robotic arm, trained to perform tasks for specific states, could successfully use our framework to diagnose, and augment, its misaligned features for new states in a new environment.

\subsection{Future work}

In this paper we took a step toward fully generalizable HRI. 
Solving model misalignment is a complex problem that can result from varied misalignment sources. While we distinguish between two -- missing and incorrect features that do not generalize to new environments -- others, such as features that have actually been learned wrongly due to limited training data, remain a question to address in future work.
In the future, we would also like to present an extended study of each of the remarks in Section \ref{ssec:framework_remarks}, and consider the possibility of allowing the robot to query the human in case there is ambiguity about which feature should be corrected.  
Another issue that would be interesting to tackle is how to distinguish when the confidence in the human input is low because the feature space is misspecified, or because the human Boltzmann observation model is wrong.

Due to lack of space, more complex examples, implementation details, and simulation videos will be linked in an extended version of this paper.

\bibliographystyle{IEEEtran}
\bibliography{bibliography_IRL}

\begin{thebibliography}{10}
\providecommand{\url}[1]{#1}
\csname url@samestyle\endcsname
\providecommand{\newblock}{\relax}
\providecommand{\bibinfo}[2]{#2}
\providecommand{\BIBentrySTDinterwordspacing}{\spaceskip=0pt\relax}
\providecommand{\BIBentryALTinterwordstretchfactor}{4}
\providecommand{\BIBentryALTinterwordspacing}{\spaceskip=\fontdimen2\font plus
\BIBentryALTinterwordstretchfactor\fontdimen3\font minus
  \fontdimen4\font\relax}
\providecommand{\BIBforeignlanguage}[2]{{%
\expandafter\ifx\csname l@#1\endcsname\relax
\typeout{** WARNING: IEEEtran.bst: No hyphenation pattern has been}%
\typeout{** loaded for the language `#1'. Using the pattern for}%
\typeout{** the default language instead.}%
\else
\language=\csname l@#1\endcsname
\fi
#2}}
\providecommand{\BIBdecl}{\relax}
\BIBdecl

\bibitem{abbeel2004apprenticeship}
P.~Abbeel and A.~Y. Ng, ``Apprenticeship learning via inverse reinforcement
  learning,'' in \emph{International Conference on Machine Learning (ICML)},
  2004.

\bibitem{bajcsy2017learning}
A.~Bajcsy, D.~P. Losey, M.~K. O’{M}alley, and A.~D. Dragan, ``Learning robot
  objectives from physical human interaction,'' in \emph{Conference on Robot
  Learning (CoRL)}, 2017, pp. 217--226.

\bibitem{christiano2017deep}
P.~F. Christiano, J.~Leike, T.~Brown, M.~Martic, S.~Legg, and D.~Amodei, ``Deep
  reinforcement learning from human preferences,'' \emph{Advances in Neural
  Information Processing Systems (NeurIPS)}, 2017.

\bibitem{javdani2015shared}
S.~Javdani, S.~S. Srinivasa, and J.~A. Bagnell, ``Shared autonomy via hindsight
  optimization,'' \emph{Robotics: Science and Systems (RSS)}, 2015.

\bibitem{ng2000algorithms}
A.~Y. Ng and S.~Russell, ``Algorithms for inverse reinforcement learning.'' in
  \emph{International Conference on Machine Learning (ICML)}, 2000, p.
  663–670.

\bibitem{ziebart2008maximum}
B.~D. Ziebart, A.~L. Maas, J.~A. Bagnell, and A.~K. Dey, ``Maximum entropy
  inverse reinforcement learning.'' in \emph{Conference on Artificial
  Intelligence (AAAI)}, 2008, pp. 1433--1438.

\bibitem{lourencco2021cooperative}
I.~Louren{\c{c}}o, R.~Mattila, C.~R. Rojas, and B.~Wahlberg, ``Cooperative
  system identification via correctional learning,'' \emph{19th IFAC Symposium
  on System Identification}, vol.~54, no.~7, pp. 19--24, 2021.

\bibitem{lourencco2022teacher}
I.~Louren{\c{c}}o, R.~Winqvist, C.~R. Rojas, and B.~Wahlberg, ``A
  teacher-student markov decision process-based framework for online
  correctional learning,'' in \emph{IEEE 61st Conference on Decision and
  Control (CDC)}, 2022, pp. 3456--3461.

\bibitem{bajcsy2018learning}
A.~Bajcsy, D.~P. Losey, M.~K. O'{M}alley, and A.~D. Dragan, ``Learning from
  physical human corrections, one feature at a time,'' in \emph{ACM/IEEE
  International Conference on Human-Robot Interaction (HRI)}, 2018, pp.
  141--149.

\bibitem{jain2015learning}
A.~Jain, S.~Sharma, T.~Joachims, and A.~Saxena, ``Learning preferences for
  manipulation tasks from online coactive feedback,'' \emph{International
  Journal of Robotics Research (IJRR)}, vol.~34, no.~10, pp. 1296--1313, 2015.

\bibitem{gutierrez2018incremental}
R.~A. Gutierrez, V.~Chu, A.~L. Thomaz, and S.~Niekum, ``Incremental task
  modification via corrective demonstrations,'' in \emph{IEEE International
  Conference on Robotics and Automation (ICRA)}, 2018, pp. 1126--1133.

\bibitem{hadfield2017inverse}
D.~Hadfield-Menell, S.~Milli, P.~Abbeel, S.~J. Russell, and A.~Dragan,
  ``Inverse reward design,'' \emph{Advances in Neural Information Processing
  Systems (NeurIPS)}, 2017.

\bibitem{ramachandran2007bayesian}
D.~Ramachandran and E.~Amir, ``Bayesian inverse reinforcement learning.'' in
  \emph{International Joint Conference on Artificial Intelligence (IJCAI)},
  vol.~7, 2007, pp. 2586--2591.

\bibitem{losey2018including}
D.~P. Losey and M.~K. O’Malley, ``Including uncertainty when learning from
  human corrections,'' in \emph{Conference on Robot Learning (CoRL)}, 2018, pp.
  123--132.

\bibitem{bobu2018learning}
A.~Bobu, A.~Bajcsy, J.~F. Fisac, and A.~D. Dragan, ``Learning under
  misspecified objective spaces,'' in \emph{Conference on Robot Learning
  (CoRL)}, 2018, pp. 796--805.

\bibitem{bobu2020quantifying}
A.~Bobu, A.~Bajcsy, J.~F. Fisac, S.~Deglurkar, and A.~D. Dragan, ``Quantifying
  hypothesis space misspecification in learning from human--robot
  demonstrations and physical corrections,'' \emph{IEEE Transactions on
  Robotics}, vol.~36, no.~3, pp. 835--854, 2020.

\bibitem{bobu2021feature}
A.~Bobu, M.~Wiggert, C.~Tomlin, and A.~D. Dragan, ``Feature expansive reward
  learning: Rethinking human input,'' in \emph{ACM/IEEE International
  Conference on Human-Robot Interaction (HRI)}, 2021, pp. 216--224.

\bibitem{weiss2016survey}
K.~Weiss, T.~M. Khoshgoftaar, and D.~Wang, ``A survey of transfer learning,''
  \emph{Journal of Big data}, vol.~3, no.~1, pp. 1--40, 2016.

\bibitem{von2007theory}
J.~Von~Neumann and O.~Morgenstern, in \emph{Theory of games and economic
  behavior}.\hskip 1em plus 0.5em minus 0.4em\relax Princeton {U}niversity
  {P}ress, 1944.

\bibitem{baker2007goal}
C.~L. Baker, J.~B. Tenenbaum, and R.~R. Saxe, ``Goal inference as inverse
  planning,'' in \emph{Proceedings of the Annual Meeting of the Cognitive
  Science Society}, vol.~29, no.~29, 2007.

\bibitem{bobu2022learning}
A.~Bobu, C.~Paxton, W.~Yang, B.~Sundaralingam, Y.-W. Chao, M.~Cakmak, and
  D.~Fox, ``Learning perceptual concepts by bootstrapping from human queries,''
  \emph{IEEE Robotics and Automation Letters}, vol.~7, no.~4, pp.
  11\,260--11\,267, 2022.

\bibitem{schulman2013finding}
J.~Schulman, J.~Ho, A.~X. Lee, I.~Awwal, H.~Bradlow, and P.~Abbeel, ``Finding
  locally optimal, collision-free trajectories with sequential convex
  optimization.'' in \emph{Robotics: Science and Systems (RSS)}, vol.~9, no.~1,
  2013, pp. 1--10.

\end{thebibliography}

\end{document}